\documentclass[10pt,twocolumn,letterpaper]{article}

\usepackage{stmaryrd}
\usepackage{iccv}
\usepackage{times}
\usepackage{epsfig}
\usepackage{graphicx}
\usepackage{amsmath}
\usepackage{amssymb}
\usepackage{amsmath}
\usepackage{amssymb}
\usepackage{dsfont}
\usepackage{bm}

\newcommand{\T}[0]{{\ensuremath{\top}}}
\newcommand{\ms}[1]{\ensuremath{\mathcal{#1}}} 
\newcommand{\mv}[1]{\ensuremath{\bm{#1}}} 
\newcommand{\mm}[1]{\ensuremath{\bm{#1}}} 

\DeclareMathOperator*{\e}{E}

\DeclareMathOperator*{\argmax}{argmax}

\newcommand{\R}[0]{\mathbb R}

\newcommand{\figref}[1]{Fig. \ref{#1}}
\newcommand{\sxnref}[1]{Section \ref{#1}}
\newcommand{\tblref}[1]{Table \ref{#1}}

\newcommand{\pgfexpand}[1]{\edef\temp{\noexpand#1}\temp} 

\usepackage{soul}
\usepackage[table]{xcolor}

\newcommand{\HJ}[1]{\hlc[yellow]{(HJ:) #1}}
\newcommand{\PP}[1]{\hlc[green]{(PP:) #1}}

\newcommand{\JZ}[1]{\hlc[pink]{(JZ:) #1}}

\def\ie{\emph{i.e.}}
\def\eg{\emph{e.g.}}

\def\acro{{\sc SuBiC}}
\def\voc{{VOC2007}}
\def\imanet{{ImageNet}}
\def\cifar{{Cifar-10}}
\def\calt{{Caltech-101}}

\newcommand{\bsm}{block-softmax}

\usepackage[mathcal]{euscript}

\def \argmax  {\arg\!\max}

\newcommand{\mbf}[1]{\ensuremath{\mathbf{#1}}}
\newcommand{\mbs}[1]{\ensuremath{\mathbf{#1}}}
\newcommand{\mcl}[1]{\ensuremath{\mathcal{#1}}}
\newcommand{\mrm}[1]{\ensuremath{\mathrm{#1}}}
\newcommand{\mbb}[1]{\ensuremath{\mathbb{#1}}}

\newcommand\mytilde[1]{\stackrel{\sim}{\smash{#1}\rule{0pt}{1.1ex}}}

\def\bI{\mbs{I}}
\def\bco{\mbs{b}}
\def\co{\mytilde{\mbs{b}}}

\def\bz{\mbs{z}}
\def\bx{\mbs{x}}
\def\barg{\operatorname{bBinEnc}_M}
\def\bsof{\operatorname{bSoftMax}_M}

\def\bb{\bco}
\def\bbm#1{\bco_{#1}}
\def\bbtbar{\overline{\bb}}
\def\bbt{\co}

\def\bzm#1{\bz_{#1}}

\usepackage{subcaption}
\usepackage{multirow}
\usepackage{authblk}

\usepackage{tabularx}
\usepackage{lipsum}

\usepackage[pagebackref=true,breaklinks=true,letterpaper=true,colorlinks,bookmarks=false]{hyperref}

\iccvfinalcopy 

\def\iccvPaperID{1413} 

\ificcvfinal\fi

\iftrue
\renewcommand{\HJ}[1]{#1}
\renewcommand{\JZ}[1]{#1}
\renewcommand{\PP}[1]{#1}
\fi

\begin{document}

\title{\textsc{SuBiC}: A supervised, structured binary code for image search}

\author[1,2]{Himalaya Jain}
\author[3]{Joaquin Zepeda}
\author[1]{Patrick P\'erez}
\author[2]{R\'emi Gribonval}
\affil[]{Technicolor, Rennes, France $\quad$ $^2$INRIA Rennes, France \quad $^3$Amazon, Seattle, WA}

\twocolumn[{%
	\renewcommand\twocolumn[1][]{#1}%
	\maketitle
	{\centering
		\vspace{-0.1cm}
		\includegraphics[width=\linewidth]{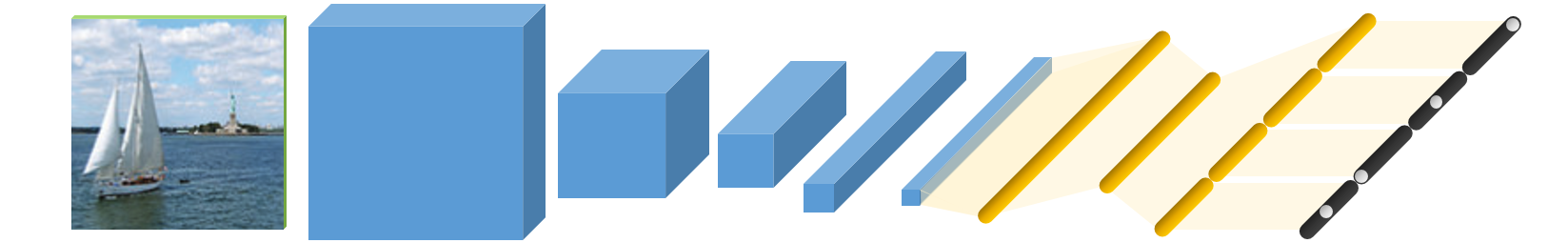}
	}
	We introduce \acro, a \textit{supervised structured binary code} (concatenation of one-hot blocks). It is produced by a novel supervised deep convolutional network and is well adapted to efficient visual search, including category retrieval for which it outperforms the state-of-the-art supervised deep binary hashing techniques.
	\vspace{0.5cm}
}]

\begin{abstract}
\JZ{ For large-scale visual search, highly compressed yet meaningful representations of images are essential. Structured vector quantizers based on product quantization and its variants are usually employed to achieve such compression while minimizing the loss of accuracy. Yet, unlike binary hashing schemes, these unsupervised methods have not yet benefited from the supervision, end-to-end learning and novel architectures ushered in by the deep learning revolution.
  We hence propose herein a novel method to  make deep convolutional neural networks produce supervised, compact, structured binary codes for visual search. Our method makes use of a novel block-softmax non-linearity and of batch-based entropy losses that together induce structure in the learned encodings. We show that our method outperforms state-of-the-art compact representations based on deep hashing or structured quantization in single and cross-domain category retrieval, instance retrieval and classification. We make our code and models publicly available online.}
\end{abstract}

\section{Introduction}\label{sec:intro}

Deep convolutional neural networks (CNNs) have proven to be versatile image representation tools with great generalization power, a quality that has rendered them indispensable in image search. A given network trained on the ImageNet dataset \cite{deng2009imagenet}, for example, can achieve excellent performance when transferred to a variety of other datasets \cite{Gong2014,kulkarni2016spleap}, or even to other visual search tasks \cite{Sharif}. This quality of \emph{transferability} is important in large-scale image search, where the time or resources to compile annotations in order to train a new network for every new dataset or task are not available.

A second desirable property of image representations for large-scale visual search is that of being \emph{compact yet functional}. A paramount example of such a representation is provided by image indexing schemes, such as Product Quantization (PQ) \cite{jegou2011product} and others, that rely on vector quantization with structured, unsupervised codebooks \cite{babenko2014additive,chen2010approximate,zhang2014composite}. PQ can be seen as mapping a feature vector into a binary vector consisting of a concatenation of one-hot encoded codeword indices. One can directly compare an uncompressed query feature with these binary vectors by means of an inner product between the binary vectors and a real-valued mapping of the query feature vector.

It is not surprising that, with the dawn of the deep learning revolution, many recent research efforts have been directed towards supervised learning of deep networks that produce compact and functional binary features \cite{dai2016binary, do2016learning,lai2015simultaneous,lin2015deep,liu2016deep,xia2014supervised,zhang2015bit,zhao2015deep}. One commonality between these approaches -- which we refer to collectively as \emph{deep hashing} methods -- is their reliance on element-wise binarization mechanisms consisting of either sigmoid/tanh non-linearities \cite{dai2016binary, lai2015simultaneous,lin2015deep,liu2016deep,xia2014supervised,zhang2015bit,zhao2015deep} or element-wise binarizing penalties such as \cite{dai2016binary, do2016learning}. Indeed, to our knowledge, ours is the first approach to impose a structure on the learned binary representation: We employ two entropy-based losses to induce a one-hot block structure in the produced  binary feature vectors, while favoring statistical uniformity in the support of the active bits of each block. The resulting structured binary code has the same structure as a PQ-encoded feature vector.



Imposing structure on the support of the binary representation has two main motivations: First, structuring allows a better exploitation of the binary representation's support to encode semantic information, \JZ{as exemplified by approaches that learn to encode face parts} \cite{bach2012structured}, visual attributes \cite{lyu2016maximum} and text topics \cite{Jenatton2010} \JZ{in the support of the representation under weak supervision.} We promote this desirable property by means of an entropy-based loss that encourages uniformity in the position of the active bits  --  a property that would not be achievable using a simple softmax non-linearity.  Note that, as a related added benefit, the structuring makes it possible to use a binary representation of larger size without incurring extra storage. Second, the structuring helps in regularizing the architecture, further contributing to increased performance relative to other, non-structured approaches.

While all previously existing deep hashing methods indeed produce very compact, functional representations, they have not been tested for transferability. The main task addressed in all these works is that of category retrieval wherein a given test example is used to rank all the test images in all classes. Yet all deep hashing approaches employ a \emph{single-domain} approach wherein the test classes and training classes are the same. It has been established experimentally \cite{sablayrolles2016should} that excellent performance can be achieved in this test by simply assigning to each stored database image, the class label produced by a classifier trained on the corresponding training set. 
Hence, it is also important to test for \emph{cross-domain} category retrieval, wherein the architecture learned on a given set of training classes is tested on a new, disjoint set of test classes. We present experiments of both types in this work, outperforming several baselines in the cross-domain test and recent deep hashing methods in the single-domain test.

The contributions of the present work can be summarized as follows:
\begin{itemize}
\item We introduce a simple, trainable, CNN layer that encodes images into structured binary codes that we coin \acro{}. While all other approaches to supervised binary encoding use element-wise binarizing operations and losses, ours are block-based.
\item We define two block-wise losses based on code entropy that can be combined with a standard classification loss to train CNNs with a \acro{} layer.
\item We demonstrate that the proposed binary features outperform the state-of-the art in single-domain category retrieval, two competitive baselines in cross-domain category retrieval and image classification, \JZ{and state-of-the art unsupervised quantizers in image retrieval.}
\item Our approach enables asymmetric search with a search complexity comparable to that of deep hashing.
\end{itemize}



\begin{table*}[h]
  \small
\caption{Comparison of proposed approach to recent supervised binary hashing techniques.}
\label{tab:synthesis}
\centering
\rowcolors{2}{gray!20}{white}
\begin{tabular}{rccccc}\rowcolor{gray!40}
Method  						& supervision 	& binarization (train -- test)&code-based loss on & base CNN training & cross-domain \\ 
CNNH+ \cite{xia2014supervised} 	& pair-wise 	& sigmoid--threshold 	& dist. to target code    & yes 			  & no\\ 
DRSCH \cite{zhang2015bit} 		& triplet-wise 	& sigmoid--threshold 	& none 					  & yes 			  & no \\  
DSRH \cite{zhao2015deep}  		& triplet-wise 		& sigmoid--threshold 	& training average 		  & yes 			  & no\\ 
DNNH \cite{lai2015simultaneous} & triplet-wise 	& sigmoid--threshold 	& none 					  & yes 			  & no\\ 
DLBHC \cite{lin2015deep}  		& point-wise 	& sigmoid--threshold 	& none 					  & fine-tuning 		  & no\\ 
DSH \cite{liu2016deep}  		& pair-wise 	& sigmoid--threshold 	& distance to binary 	  & yes 			  & no\\ 
BDNN \cite{do2016learning}  	& pair-wise 	& built-in 				& feature reconst. error  & no 				  & no\\ 
\acro~(ours)  					& point-wise 	& block softmax--argmax & block-based entropies   & yes	 		 	  & yes 
\end{tabular}
\normalsize
\end{table*}

\section{Related work}\label{sec:related}

We discuss here the forms of vector quantization and binary hashing that are the most important for efficient visual search with compact codes, and we explain how our approach relates to them.       

\paragraph{Unsupervised structured quantization.} Vector quantization (VQ), \eg~with unsupervised $k$-means, is a classic technique to index multi-dimensional data collections in a compact way while allowing efficient (approximate) search. Structured versions of VQ, \eg~product, additive or composite \cite{babenko2014additive,ge2013optimized,ge2014product,jain2016approximate,jegou2011product,kalantidis2014locally,nourouzi2013cartesian,zhang2014composite}, have established impressive indexing systems for large scale image collections. Coupled with single or multiple index inverted file systems \cite{babenko12inverted,jegou2011searching}, these VQ techniques currently offer state-of-the-art performance for very large-scale high-dimensional nearest-neighbor search (relative to the Euclidean distance in input feature space) and instance image search based on visual similarity. All these unsupervised quantization techniques operate on engineered or pre-trained image features. Owing to the success of CNNs for image analysis at large, most recent variants use off-the-shelf or specific CNN features as input representation, \eg, \cite{babenko2016efficient,johnson2017billion,liu2015indexing,wang2016face}. However, contrary to binary hashing methods discussed below, VQ-based indexing has not yet been approached from a supervised angle where available semantic knowledge would help optimize the indexed codes and possibly the input features. In the present work, we take a supervised encoding approach that bears a strong connection to supervised binary hashing, while exploiting an important aspect of these powerful unsupervised VQ techniques, namely the structure of the code. The binary codes produced by our approach are in a discrete product space of size $K^M$ while allowing $\mathcal{O}(M\log K)$ storage and $\mathcal{O}(M)$ search complexity.

\paragraph{Deep, supervised hashing.} Binary hashing is a long-standing alternative to the above-mentioned VQ methods,
and the deep learning revolution has pushed the state-of-the-art of these approaches.  
Deep supervised hashing methods --  
CNNH+ \cite{xia2014supervised}, 
DRSCH \cite{zhang2015bit}, 
DSRH \cite{zhao2015deep}, 
DNNH \cite{lai2015simultaneous}, 
DLBHC \cite{lin2015deep}, 
DSH \cite{liu2016deep} and 
BDNN \cite{do2016learning} 
-- share the following high-level principles. 
An off-the-shelf or home-brewed convolutional network $f_1$ is used to extract a high-dimensional distributed representation $\bx\in\mathbb{R}^d$ from an input image $\bI$.
A subsequent fully connected encoding layer $f_2$ turns this feature vector into a compact binary code $\mbs{h}\in\{0,1\}^B$ of $B$ bits through final entry-wise thresholding (or sign function for centered codes), $B$ typically ranging from 12 to 64 bits.
At training time, this binarization is usually relaxed using a sigmoid (or $\tanh$ for centered codes) -- with the exception of BDNN \cite{do2016learning} --, which results in an encoding layer that outputs vectors in $\in [0,1]^B$. Using semantic supervision, $f_2$ is trained while $f_1$ is fixed to pre-trained values, fine-tuned or trained from scratch. Supervision coming from class labels is used either directly (classification training) \cite{lin2015deep} or using tuples (pairs \cite{do2016learning,liu2016deep,xia2014supervised} or triplets \cite{lai2015simultaneous,zhang2015bit,zhao2015deep}) as in metric learning. Table \ref{tab:synthesis} summarizes the specifics of each of these methods.   

DLBHC \cite{lin2015deep} is a simple instance employing a sigmoid-activated encoding layer grafted to the pre-trained AlexNet architecture \cite{krizhevsky2012imagenet} and trained using a standard classification objective. 
Other methods employ additional training loss(es) at the code level to induce desirable properties. BDNN \cite{do2016learning} uses the code-to-feature reconstruction error, making the approach applicable in an unsupervised regime. DSH \cite{liu2016deep} employs a W-shaped loss with minima at the desired code values, while DSRH \cite{zhao2015deep} penalizes the average of each bit over the training set such that its distribution is approximately centred (final code is in $\{-1,+1\}^B$). CNN+ \cite{xia2014supervised} proposes the direct supervision of hash functions with target binary codes learned in a preliminary phase  via low-rank factorization of a full pairwise similarity matrix. Note also that DRSCH \cite{zhang2015bit} learns bit-wise weights along with the binary encoder, which results in richer codes but more costly distances to compute at search time.


As described in detail next, our approach follows the same high-level principles discussed above, but with important differences. The first, key difference lies in the structure of the codes. 
We define them as the concatenation of $M$ one-hot vectors (binary vectors with all entries but one being zero) of size $K$. This gives access to $K^M$ distinct codes, hence corresponding to an effective bit-size of $M\log_2 K$ bits, as in VQ methods that combine $M$ codewords, each taken from a codebook of size $K$. Binarization and its relaxed training version thus operate at the block level. 
This specific code structure is combined with novel loss terms that enforce respectively the one-sparsity of each block and the effective use of the entire block support. Also, contrary to most supervised binary hashing approaches, with the exception of \cite{lin2015deep}, we resort only to point-wise supervision.

\JZ{}\HJ{To our knowledge, only one other approach has incorporated product-wise structuring within a deep learning pipeline} \cite{cao2016deep}. \HJ{Yet that method does not learn the structuring as part of a deep architecture, relying rather on a standard product quantizer that is updated once per epoch in an unsupervised manner.}




\section{Approach}\label{sec:approach}

We describe in this section the design of our \acro{} architecture, its supervised training and its use for visual search.

\begin{figure*}
  \centering
  
  \includegraphics[scale=0.480]{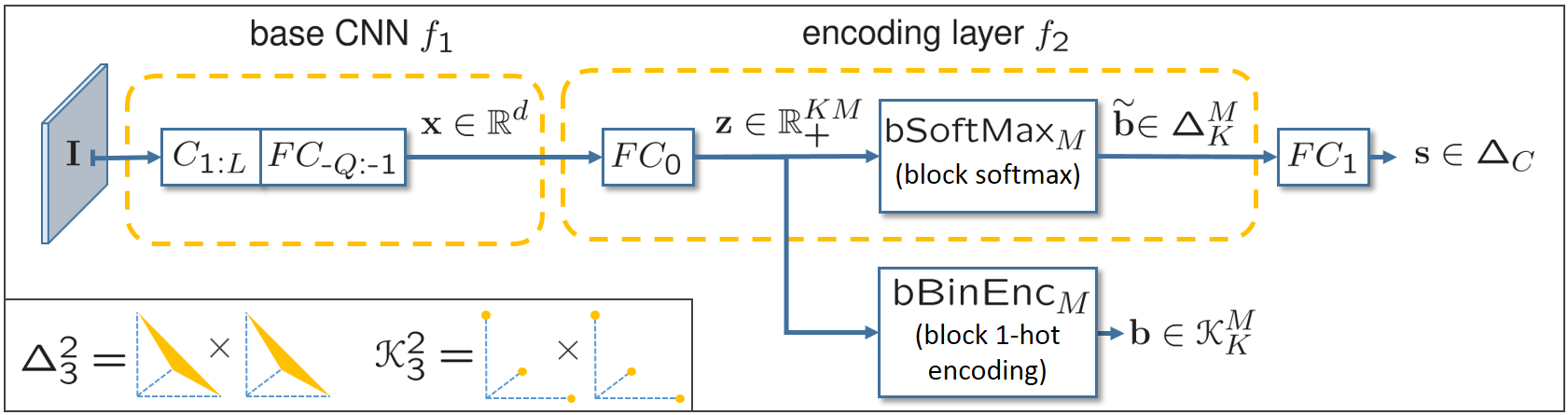}
  \caption{{\bf Proposed architecture and notations}. \JZ{A feature is extracted from image $\bI$ by a base CNN $f_1$ and binarized using a block-structured encoding layer $f_2$ consisting of a fully-connected layer followed by a \emph{block softmax} during training, or a \emph{block $1$-hot encoder} during testing.}}
  \label{fig:net}
\end{figure*}

\subsection{\JZ{Architecture}} 


Following the approach discussed above, we consider the following classification feed-forward network (Fig. \ref{fig:net}): 
\begin{equation}
  \label{eq:s}
  \mbf{s}  \triangleq FC_1 \circ f_2 \circ f_1 ( \bI ),
\end{equation}
where $\bI$ is an input image, $f_1$ a deep CNN with $L$ convolutional layers (inc. pooling and normalization, if any) and $Q$ fully-connected layers, $f_2$ a binary encoding layer, $FC_1$ a $C$-class classification layer, and $\mbf{s}$ the $C$-dimensional vector of class-probability estimates.




We aim for the \JZ{binary encoding layer $f_2$} to produce structured binary vectors $\bco$ consisting of the concatenation of $M$ one-hot encoded vectors $\bco_m, m=1,\ldots,M,$ of dimension $K$, \ie, $\bco = [\bco_1; \ldots; \bco_M]$.\footnote{Using vector stacking notation $[\mbf{a};\mbf{b}] = [\mbs{a}^\top,\mbf{b}^\top]^\top$, where $\mbf{a}$ and $\mbf{b}$ are column-vectors.} Formally, the \emph{blocks} $\bco_m$ should satisfy
\begin{equation} \label{eq:constraint}
  \bco_m \in \ms K_K \triangleq \{ \mbf{d} \in \{0,1\}^K \textrm{ s.t. } \|\mbf{d}\|_1=1 \}.
\end{equation}
Accordingly, our codes $\bco$ should come from the discrete product set $\mathcal{K}_K^M$.

In practice, $f_2$  employs a fully-connected layer $FC_0$ with \JZ{ReLU non-linearity} producing real-valued vectors $\bz \in \R_+^{KM}$ likewise consisting of $M$ $K$-dimensional blocks $\bz_m$. A second non-linearity operates on each $\bz_m$ to produce the corresponding binarized block. We use a different binarization strategy at training time (top branch in \figref{fig:net}) and at test time (bottom branch), as discussed next.


\paragraph{Training architecture.} Similarly to supervised binary hashing approaches discussed in Section \ref{sec:related}, we enable back-propagation during our learning process by relaxing the structured binarization constraint \eqref{eq:constraint}, producing instead structured real-valued codes $\co \,\,\in \Delta_K^M$, where
\begin{equation}
  \Delta_K \triangleq \{\mbf{d} \in [0,1]^K\textrm{ s.t. }  \|\mbf{d}\|_1 = 1\}
\end{equation}
\JZ{is the convex hull of $\mathcal K_K$} (see \figref{fig:net} bottom-left, for the examples $\Delta_3^2$ and $\mathcal{K}_3^2$). We achieve this by \JZ{introducing} the \textit{block-softmax non-linearity}  $\co = \bsof(\bz)$ (\cf Fig. \ref{fig:net}) which computes the blocks $\co_m$ from the corresponding blocks $\bz_m$ as follows ($\exp(\cdot)$ denotes element-wise exponentiation):
\begin{equation}
  \co_m \,\, = \frac{1}{\|\exp (\bz_m)\|_1} \exp ( \bz_m).
\end{equation}



\paragraph{Test time architecture.} 
At test time, the block softmax is replaced (\textit{cf.} \figref{fig:net}, bottom branch) by a \textit{block one-hot encoder} $\bco = \barg(\bz)$, which uses $\bz$ to efficiently compute the projection of each block $\co_m \,\,\in \Delta_K$ onto $\mathcal{K}_K$ using
\begin{equation}
  \bco_m[k] =
  \begin{cases}
    1 & \textrm{ if } k = \argmax_r \bz_m[r], \\
    0 & \textrm{ otherwise,} 
  \end{cases}
\label{eq:bargmin2}
\end{equation}
where $\mbf{d}[k]$ denotes the $k$-th entry of a vector $\mbf{d}$. Note, particularly, that $\barg(\bz) = \barg(\co)$. 

\subsection{Supervised loss and training}\label{sec:training}
In order to bring real-valued code vectors $\co$ as close as possible to block-wise one-hot vectors, while making the best use of coding budget,  
we introduce two entropy-based losses that will be part of our learning objective. Our approach assumes a standard learning method wherein training examples $(\bI^{(i)}, y^{(i)})$ consisting of an image $\bI^{(i)}$ and its class label $y^{(i)} \in \{1, \ldots, C\}$ are divided into mini-batches $\{(\bI^{(i)}, y^{(i)})\}_{i \in \ms T}$ of size $|\ms T|=T$. 

Our losses will be based on  \emph{entropy}, which is computed for a vector $\mbf{p} \in \Delta_K$ 
as follows:\footnote{With the usual convention $0\log_2(0)=0$.}
\begin{equation} \label{eq:entropy}
  \e(\mbf{p}) \triangleq - \sum_{k=1}^K \mbf{p}[k] \log_2 \mbf{p}[k].
\end{equation}
Entropy is smooth and convex and further has the interesting property that it is the theoretical minimum average number of bits per symbol required to encode an infinite sequence of symbols with distribution $\mbf{p}$ \cite{cover2006elements}.  Accordingly, it is exactly zero, its minimum, if $\mbf{p}$ specifies a deterministic distribution (\ie, $\mbf{p} \in \mathcal K_K$) and $\log_2 K$, its maximum, if it specifies a uniform distribution (\ie, $\mbf{p} = \frac{1}{K} \mbf{1}$). 

\paragraph{Toward block-wise one-hot encoding.} Given the merits of structured binary codes discussed previously, we aim to produce feature vectors $\co\,\, =[ \co_1; \ldots; \co_M ]$ consisting of blocks $\co_m$ that approximate one-hot encoded vectors, thus that have a small projection error
\begin{equation}\label{eq:proj err}
  \min_{\mbf{d} \in \ms K_K } \| \mbf{d} - \co_m \!\!\|_2.
\end{equation}

In the ideal case where $\co_m \,\,\in \ms K_K$, it has minimum entropy of $0$. The convexity/smoothness of $\e(\cdot)$ means that blocks with low entropy will have small projection error \eqref{eq:proj err}, thus suggesting penalizing our learning objective for a given training image using  $\sum_m \e(\co_m)$. We overload our definition of $\e(\cdot)$ in \eqref{eq:entropy} and let $\e(\co) \triangleq \sum_m \e(\co_m) \in [0, M\log_2 K]$. Accordingly, we refer to the average of these losses over a training batch $\ms T$ as the \emph{mean entropy}, given by
\begin{equation}
\frac{1}{TM} \sum_{i \in \ms T} \sum_{m=1}^M \e(\co_m^{(i)}) = \frac{1}{TM} \sum_{i \in \ms T} \e(\bbt^{(i)}).
\label{eq:mean entropy}
\end{equation}
In practice, introducing this loss will result in vectors $\co$ that are only approximately binary, and hence, at test time, we project each block $\co_m$ onto $\ms K_K$ using \eqref{eq:bargmin2}.


\paragraph{Uniform block support.} Besides having blocks $\bbt_m$ that resemble one-hot vectors, we would like for the supports of the binarized version $\bco_m$ of $\co_m$ to be as close to uniformly distributed as possible. This property allows the system to better exploit the support of our $\bbt$ in encoding semantic information. It further contributes to the regularization of the model and encourages a better use of the available bit-rate.

We note first that one can estimate the distribution of the support of the $\bco_m$ from a batch $\ms T$ using
$\frac{1}{T} \sum_{i \in \ms T} \bco_m^{(i)}$. Relaxing $\bco_m$ to $\co_m$ for training purposes, we want the entropy of this quantity to be high. 
This leads us to the definition of the negative \emph{batch entropy} loss: 
\begin{equation}\label{eq:batch entropy}
  -\frac{1}{M}\sum_{m=1}^M \e \Big(\frac{1}{T}\sum_{i \in \ms T} \co_m^{(i)} \Big ) = - \frac{1}{M} \e \left ( \bbtbar \right ),
\end{equation}
where we let
$
\bbtbar\triangleq \frac{1}{T}\sum_{i \in \ms T} \bbt^{(i)}
$.

Our learning objective (computed over a mini-batch) will hence be a standard classification objective further penalized by the mean and batch entropies in \eqref{eq:mean entropy} and \eqref{eq:batch entropy}:
\begin{equation}\label{eq:batch loss}
\begin{split}
\mathrm{Loss}\big(\{(\bI^{(i)},& y^{(i)})\}_{i\in \ms T}\big) \triangleq  \frac{1}{T}   \sum_{i \in \ms T} \Big[ \ell(\mbf{s}^{(i)}, y^{(i)}) +  \\ 
&  \frac{\gamma}{M \log_2 K}\e(\bbt^{(i)})  - \frac{\mu}{M \log_2 K} \e (\bbtbar ) \Big],
\end{split}
\end{equation}
with network output $\mbf{s}$ defined as in (\ref{eq:s}), $C$ the number of classes, and $\gamma>0$ and $\mu>0$ two hyper-parameters. 

In our work, we use the following scaled version of the commonly used cross-entropy loss for classification: 
\begin{equation}
  \ell(\mbf{s}, y) \triangleq - \frac{1}{\log_2 C} \log_2 \mbf{s}[y].
\end{equation}
The scaling by $\log_2C$ reduces the dependence of the hyper-parameters $\mu$ and $\gamma$ on the number of classes $C$.

The training loss (\ref{eq:batch loss}) is minimized with mini-batch stochastic gradient descent. The whole architecture can be learned this way, including the CNN feature extractor, the encoding layer and the classification layer (Fig. \ref{fig:net}). Alternatively, (some of) the weights of the base CNN $f_1$ can be fixed to pre-trained values. In Section \ref{sec:experiments}, we will consider the following variants, depending on set-ups: Training of $FC_0 / FC_1$ only (``2-layer'' training), the base CNN staying fixed; Training of $FC_{-1}/FC_0/FC_1$ (``3-layer training''); Training of all layers, $C_1\cdots C_L$ and $FC_{-Q}\cdots FC_1$ (``full training'').  

\subsection{Image search}
\label{sec:tasks}
As we will establish in Section \ref{sec:experiments}, \acro{} yields important advantages in \JZ{three} image search applications, which we now describe \JZ{along with a search complexity analysis.}

\paragraph{Category and instance retrieval.} These two tasks consist of \JZ{ranking database images according to their similarity to a given query image, where similarity is defined by membership in a given semantic category \textit{(category retrieval)} or by the presence of a specific object or scene \textit{(instance retrieval)}. For these two tasks,} we wish to use our structured binary representations to efficiently compute similarity scores for all database of images $\{\bI^{(j)}\}_j$ given a query image $\bI^*$. We propose using an asymmetric approach \cite{jegou2011product} that limits query-side coding approximation: The database images are represented using their structured binary representation $\bb^{(j)} = [\bb^{(j)}_1;\cdots;\bb^{(j)}_M] \in \ms K_K^M$, whereas the query image $\bI^*$ is represented using the real-valued vector $\bz^* = [\bz^*_1;\cdots;\bz^*_M] \in \R_+^{KM}$. Accordingly, the database images are ranked using the similarity score $(\bz^*)^{\top} \bb^{(j)}$. This expression also reads
\begin{equation} \label{eq:LUT}
  \sum_{m=1}^M \bzm{m}^* \big[ \argmax_r \bco_m^{(j)}[r] \big],
\end{equation}
which shows that $M$ additions are needed to compute \acro{} similarities.


\paragraph{Image classification.} A second important application is that of image classification in the case where the classes of interest are not known beforehand or change across time, as is the case of on-the-fly image classification from text queries \cite{chatfield2012visor,chatfield2014efficient}. Having feature representations that are compact yet discriminative is important in this scenario, and a common approach to achieve this is to compress the feature vectors using PQ \cite{chatfield2015fly,chatfield2014efficient}. The approach we propose is to instead use our supervised features to compactly represent the database images directly. New classes are assumed to be provided in the form of annotated sets $\{(\mbf{I}^{(q)}, y^{(q)})\}_q$ containing examples of the $C^\prime$ previously-unknown query classes,\footnote{Obtained from an external image search engine in on-the-fly scenarios.} and
we learn classifiers from the structured codes $\bbt$ of these examples.
At test time, classifying a test feature $\bco^{(j)} \in \ms K_{K}^M$ from the original dataset will require computing products $(\mm W^*)^\T \bco^{(j)}$ (with $\mm W^{*} \in \R^{KM \times C^\prime }$ for a softmax classifier or $\mm W^* \in \R^{KM}$ for a one-vs-rest classifier). Similarly to \eqref{eq:LUT}, this operation will likewise require only $M$ additions per column of $\mm W^*$.


\paragraph{Search complexity relative to deep hashing.} 
The expression \eqref{eq:LUT} is reminiscent of the efficient distance computation mechanisms based on look-up-tables commonly used in product quantization search methods \cite{jegou2011product}. 
In particular, the expression in \eqref{eq:LUT} establishes that computing the similarity between $\bzm{m}^*$ and $\bco_m$ incurs a complexity of $M$ additions. This can be compared to the complexity incurred when computing the Hamming distance between two deep hash codes (\textit{cf.} \sxnref{sec:related}) $\mbf{h}_1$ and $\mbf{h}_2$ of length $B=M \log_2 K$ (\ie, of storage footprint $B$ equal to that of \acro{}): $1$ XOR operation followed by 
as many additions as there are different bits in $\mbf{h}_1$ and $\mbf{h}_2$, a value that can be estimated from the expectation ($\llbracket \cdot \rrbracket$ is the Iverson bracket)

\begin{equation}
  {\mathbb{E}}_{\mbf{h}_1, \mbf{h}_2} \left ( \sum_{k=1}^{B} \big\llbracket \mbf{h}_1[k] \neq \mbf{h}_2[k] \big\rrbracket  \right )= \frac{B}{2} = \frac{M}{2} \log_2 K,
\end{equation}
if assuming i.i.d. and uniform $\mbf{h}_j[k]$.

We note that $\mathcal{O}(1)$  look-up-table (LUT) based implementations of the Hamming distance are indeed possible, but only for small $B$ (the required LUT size is $2^B$). Alternatively, a smaller LUT of size $2^{B/M^\prime}$ can be used by splitting the code into $M^\prime$ blocks (with $M^\prime$ comparable to $M$), resulting in a complexity $\mathcal{O}(M^\prime)$ comparable to the $\mathcal{O}(M)$ complexity of \acro{}.



\def\acro{{\sc SuBiC}}
\section{Experiments}\label{sec:experiments}

We assess the merits of the proposed supervised structured binary encoding for \HJ{instance and} semantic image retrieval by example and for database image classification, the three tasks described in Section \ref{sec:tasks}.

\paragraph{Single-domain category retrieval.}
Single-domain category retrieval is the main experimental benchmark in the supervised binary hashing literature.
Following the experimental protocol of \cite{liu2016deep}, we report mean average precision (mAP) performance on the \cifar~database \cite{cifar_url} which has 10 categories and 60k images of size $32\times 32$ for each. The training is done on the 50k image training set. The test set is split into 9k database images and 1k query images, 100 per class. For fairness of comparison, we also use as base CNN the same as introduced in \cite{liu2016deep}. It is composed of $L=3$ convolutional layers with $32$, $32$ and $64$ filters of size $5\times5$ respectively, followed by a fully connected layer $FC_{-1}$ with $d=500$ nodes. As proposed, we append to it a randomly initialized \textit{encoder layer} $FC_0$ along with the classification layer $FC_1$. We fixed $K=64$ and varied $M=\{2,4,6,8\}$ so that $B=M\log_2(K)$ is equal to the desired bit-rate. Full training of the network is conducted, and hyper-parameters $\gamma$ and $\mu$ are cross-validated as discussed later. We compare in \tblref{tab:cifar10} with various methods based on the same base CNN (top four rows, 
DSH \cite{liu2016deep}, DNNH \cite{lai2015simultaneous}, DLBHC \cite{lin2015deep} and CNNH+ \cite{xia2014supervised}), as well as other published values. For reference, we include a method (KSH-CNN \cite{liu2012supervised}) not based on neural hash functions but using activations of a deep CNN as input features.
Note that, at all bit-rates, from 12 to 48 bits, \acro~outperforms these state-of-the-art supervised hashing techniques. 


\begin{table}[!t]
  \small
\centering
\rowcolors{2}{gray!20}{white}
  \begin{tabular}{rcccc}  \rowcolor{gray!40}
	Method & 12-bit & 24-bit & 36-bit & 48-bit \\ 
	CNNH+ \cite{xia2014supervised} & 0.5425 & 0.5604 & 0.5640 & 0.5574 \\ 
	DLBHC \cite{lin2015deep} & 0.5503 & 0.5803 & 0.5778 & 0.5885 \\ 
	DNNH \cite{lai2015simultaneous} & 0.5708 & 0.5875 & 0.5899 & 0.5904 \\ 
	DSH \cite{liu2016deep} & 0.6157 & 0.6512 & 0.6607 & 0.6755 \\ 
    KSH-CNN \cite{liu2012supervised}&-& 0.4298&-			& 0.4577 \\
	DSRH \cite{zhao2015deep} 		&-& 0.6108&-			& 0.6177 \\ 
	DRSCH \cite{zhang2015bit} 		&-& 0.6219&-			& 0.6305 \\ 
	BDNN \cite{do2016learning} 		&-& 0.6521&-			& 0.6653 \\ 
        \acro~(ours) & \textbf{0.6349} & \textbf{0.6719} & \textbf{0.6823} & \textbf{0.6863} \\
  \end{tabular} \\
  \caption{\textbf{Single-domain category retrieval.} Comparison against published mAP values on \cifar\ for various supervised deep hashing methods. \JZ{See the} \emph{\imanet} column of \tblref{tab:eval_category} for single-domain results on \imanet.}
  \label{tab:cifar10}
\end{table}
\begin{table}[!t]
  \small
\centering
	\rowcolors{2}{gray!20}{white}
	\begin{tabular}{rccc} \rowcolor{gray!40}
		Method & \voc & \calt & \imanet \\ 
		PQ \cite{jegou2011product} & 0.4965 & 0.3089 & 0.1650 \\ 
		CKM \cite{nourouzi2013cartesian} & 0.4995 & 0.3179 & 0.1737\\
		\HJ{LSQ} \cite{martinez2016revisiting} & 0.4993 & 0.3372 & 0.1882\\
		\HJ{DSH-64} \cite{liu2016deep} & 0.4914 & 0.2852 & 0.1665\\
		\acro~2-layer & \textbf{0.5600} & 0.3923 & 0.2543 \\ 
		\acro~3-layer & 0.5588 & \textbf{0.4033} & \textbf{0.2810} \\ 
    \end{tabular} \\
	\caption{\textbf{Cross-domain category retrieval.} Performance (mAP) using 64-bit encoders across three different datasets using VGG-128 as base feature extractor. For completeness, results on \imanet{} validation set (\ie\ single-domain retrieval) are provided in the third column. }\label{tab:eval_category}
\end{table}


\paragraph{Cross-domain category retrieval.}
Using VGG-D with $128$-D bottleneck (VGG-128) \cite{chatfield2014return} as base CNN ($L=5$, $Q=3$ and $d=128$), setting $\mu$ and $\gamma$ to $1.0$, we performed 2-layer and 3-layer learning of our network (see Section \ref{sec:training}) on ILSVRC-\imanet\ \cite{ilsvrc_url} training set. Two-layer training is conducted on 5k batches of $T=200$ images. Three-layer training is initialized by previous one and run for another 5k batches. To evaluate cross-domain performance, we used our trained network to do category retrieval on Pascal \voc \cite{voc_url}, \calt\ \cite{caltech_url} and \imanet~validation sets. For each experiment, we used 1000 (2000 for \imanet) random query images, the rest serving as database. The performance of the two trained \acro~networks is reported in Table \ref{tab:eval_category} at 64-bit rate ($M=8$, $K=256$). They are compared to three unsupervised quantization baselines, PQ \cite{jegou2011product},  Cartesian \textit{k}-means (CKM) \cite{nourouzi2013cartesian} \HJ{and LSQ} \cite{martinez2016revisiting}, operating at 64-bit rate on VGG-128 image features. \HJ{Further, to compare with supervised deep hashing approaches we implemented DSH} \cite{liu2016deep} \HJ{with VGG-128 as the base CNN, using their proposed loss and pair-wise training.}

The impact of the proposed semantic supervision across domain is clearly demonstrated. Comparing unsupervised methods with our ``2-layer'' trained variant (no tuning of $FC_{1}$ is particularly enlightening since they all share exactly the same 128-dimensional input features). Training this representation as well in the ``3-layer'' version did not prove useful except on the \imanet~validation set. Note that the performance on this set could have been further improved through longer training, but at the expense on reduced transferability.

\paragraph{\JZ{}\HJ{Instance retrieval.}} Unsupervised structured quantizers produce compact codes that enjoy state-of-the-art performance in instance retrieval at low memory footprint. Hence, in \tblref{tab:eval_instance} we compare \acro{} to various such quantizers as well as DSH, \JZ{using $64$-bit representations for all methods}. We used the \textit{clean train} subset \cite{gordo2016deep} of the Landmarks dataset \cite{babenko2014neural} to train both DSH and a $2$-layer \acro{} (the same as in \tblref{tab:eval_category}, but with 60K batches). We report mAP on the Oxford 5K \cite{philbin2007object} and Paris 6K \cite{philbin2008lost} datasets using their provided query/database split. \acro~ outperforms all methods while DSH performance is weaker to even unsupervised quantizers.


\begin{table}[!t]
  \small
	\centering
	\rowcolors{2}{gray!20}{white}
	\begin{tabular}{rcc} \rowcolor{gray!40}
		Method & Oxford5K & Paris6K \\ 
		PQ \cite{jegou2011product} & 0.2374 & 0.3597 \\ 
		LSQ \cite{martinez2016revisiting} & 0.2512 & 0.3764 \\
		DSH-64 \cite{liu2016deep} & 0.2108 & 0.3287 \\
		\acro & \textbf{0.2626} & \textbf{0.4116} \\
	\end{tabular} \\
	\caption{\HJ{\textbf{Instance retrieval.}} \JZ{Performance (mAP) comparison using $64$-bit codes for all methods.} }\label{tab:eval_instance}
\end{table}

\paragraph{Image classification.}
In \tblref{tab:classification}, we show how the 64-bit \acro~encoding of VGG-128 features from \tblref{tab:eval_category} (2-layer variant) outperforms two baseline encoders with the same bit-rate for classification of compressed representations. 
As done in \cite{chatfield2014efficient} for on-the-fly classification, the first baseline employs PQ \cite{jegou2011product} to represent the features compactly, and the second substitutes PQ by the better-performing CKM encoder \cite{nourouzi2013cartesian}. Both unsupervised encoders are learned on VGG-128 features from the \imanet~training set. For the test on \imanet, the two baselines employ the off-the-shelf VGG-128 classification layer as a classifier, reconstructing the PQ and CKM encoded versions beforehand (for reference, first row is for this classifier using original, un-coded features). 
Our results (bottom two rows) employ trained $FC_1$ layer applied to either $\bbt$ code (``\acro{} soft'') or $\bb$ binary code (``\acro{} $64$-bit''). In case of \voc~we trained one-vs-rest SVM classifiers on the off-the-shelf VGG-128 features (top three rows) or on the $\bbt$ features for \acro~(bottom two rows).

Note that our compact \acro~64-bit features outperform both PQ and CKM features for the same bit-rate. Also notice that, although the classifiers for \voc\ are trained on block-softmax encoded features, when we use \acro~64-bit features the accuracy drops only marginally.

\begin{table}[!t]
  \small
\centering
\rowcolors{2}{gray!20}{white}
	\begin{tabular}{rccc} \rowcolor{gray!40}
           & \multicolumn{2}{c}{\imanet} & \voc  \\
          \rowcolor{gray!40}   & \emph{Top-1 acc.} & \emph{Top-5 acc.} & \emph{mAP} \\ 
		VGG-128$^*$& 53.80 & 77.32 & 73.79 \\ 
		PQ 64-bit~ & 39.88 & 67.22 & 65.94 \\ 
		CKM 64-bit~ & 41.15 & 69.66 & 67.25 \\ 
		\acro~soft$^*$ & 50.07 &  74.11 & 70.20 \\ 
		\acro~64-bit~ & 47.77 & 72.16 & 67.86 \\ 
	\end{tabular} \\
	\caption{\textbf{Classification performance with different compact codes}.
          The rows marked (*) are non-binary codes. See the text for details.}\label{tab:classification}
\end{table}


\begin{figure}[!t]
  \centering
  \input{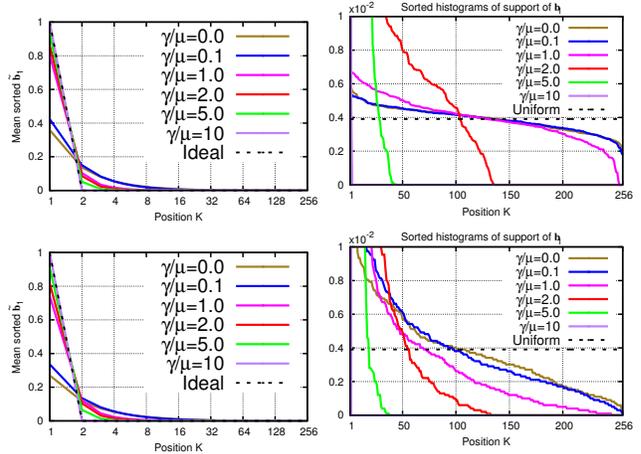}
  \caption{\textbf{Effect of the entropy-based losses on the behavior of structured encoding.} (\textit{left}) One-hot encoding closeness of $\co_1$. (\textit{right}) Distribution of block support of $\bbm{1}$. The black dashed curves correspond to the ideal, desired behavior.  (\textit{top}) \imanet~validation. (\textit{bottom}) \voc.   
  }
  \label{fig:losses}
\end{figure}
\begin{figure}[!t]
  \includegraphics[width=\linewidth, height=1.5in]{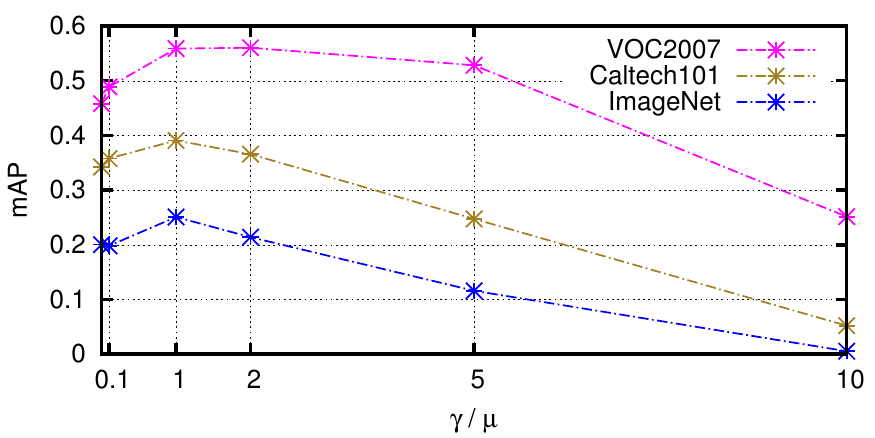}
  \caption{\textbf{Effect of $\gamma$ and $\mu$ on category retrieval performance.} 
    }
  \label{fig:xval}
\end{figure}

\paragraph{\JZ{Structuring effectiveness of entropy-based losses.}} 
In \figref{fig:losses} we evaluate our proposed entropy losses using the same \acro{} setup as in \tblref{tab:classification}. We report statistics on the \imanet~validation set (top graphs) and on all of \voc~(bottom graphs), using the $\gamma/\mu$ ratio in the legends for ease of comparison.

To explore how well our $\gamma$-weighted mean-entropy loss favors codes resembling one-hot vectors, we extract the first 256-dimensional block $\co_1$ from each image of the set (the seven other blocks exhibit similar behavior), re-order the entries of each such $\co_1$ in decreasing order and average the resulting collection of vectors. The entries of this average vector are visualized for various values of $\gamma/\mu$ in the plots on the left. Ideal one-hot behavior corresponds to $[1;0\cdots;0]^\top$. On both datasets, increasing the mean entropy penalization weight $\gamma$ relative to $\mu$ (\ie, increasing $\gamma/\mu$) results in code blocks that more closely resemble one-hot vectors.





To evaluate how well our $\mu$-weighted negative batch-entropy term promotes uniformity of the support of the binarized blocks in $\bco$, we plot, in the right side of \figref{fig:losses}, the sorted histograms of the support of the first block $\bbm{1}$ over the considered image set.
Note that increasing the weight $\mu$ of the batch entropy term relative to $\gamma$ (decreasing $\gamma/\mu$) results in distributions that are closer to uniform. As expected, the effect is more pronounced on the \imanet~dataset (top row) used as a training set, but extrapolates well to an independent dataset (bottom row).

We note further that it is possible (green curves, $\gamma/\mu=5$) to have blocks that closely resemble one-hot vectors (left plot) but make poor use of the available support ($0$-valued histogram after the $47$-th bin, on the right). It is likewise possible (blue curve, $\gamma/\mu=0.1$) to enjoy good support usage with blocks that do not resemble one-hot vectors, establishing that our two losses work together to achieve the desired design goals.
 


\paragraph{\JZ{Cross-validation of hyper parameters.}} Using the same setup and $\gamma/\mu$ values as in \figref{fig:losses}, in \figref{fig:xval} we plot mAP as a function of $\gamma/\mu$ on three datasets. Note that the optimal performance (at $\gamma/\mu=1$) for this architecture is obtained for an operating point that makes better use (closer to uniform) of the support of the blocks, as exemplified by the corresponding curves (pink) on the right in \figref{fig:losses}. This supports one of our original motivations that fostering uniformity of the support would encourage the system to use the support to encode semantic information.

\begin{figure}[!t]
  \centering
  \xdef\tempwidth{0.8\linewidth}
\def\tmpscale{1.08}
\begin{tabularx}{\linewidth}{c  X}    
  \multirow{2}{*}{\includegraphics[scale=\tmpscale]{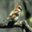}} &
  \multicolumn{1}{m{\tempwidth}}{
    \includegraphics[scale=\tmpscale]{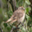}
    \includegraphics[scale=\tmpscale]{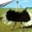}
    \includegraphics[scale=\tmpscale]{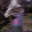}  
    \includegraphics[scale=\tmpscale]{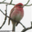}
    \includegraphics[scale=\tmpscale]{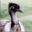}\newline%
    \includegraphics[scale=\tmpscale]{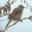}
    \includegraphics[scale=\tmpscale]{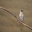}
    \includegraphics[scale=\tmpscale]{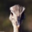}
    \includegraphics[scale=\tmpscale]{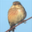}  
    \includegraphics[scale=\tmpscale]{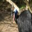}\newline
  } \\
  \vspace{1em}
  & \multicolumn{1}{m{\tempwidth}}{
    \includegraphics[scale=\tmpscale]{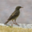}
    \includegraphics[scale=\tmpscale]{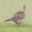}
    \includegraphics[scale=\tmpscale]{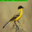}
    \includegraphics[scale=\tmpscale]{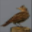}
    \includegraphics[scale=\tmpscale]{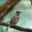}\newline%
    \includegraphics[scale=\tmpscale]{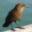}
    \includegraphics[scale=\tmpscale]{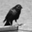}
    \includegraphics[scale=\tmpscale]{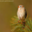}
    \includegraphics[scale=\tmpscale]{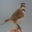}
    \includegraphics[scale=\tmpscale]{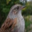}}
  \\ 
\end{tabularx}

  \caption{\textbf{Category retrieval examples}. Top ten ranked images retrieved from \cifar~for the query on the left when using $12$-bit (\emph{top}) and $48$-bit (\emph{bottom}) \acro. Note that higher bit-rates make the representation more sensitive to  the query's orientation. 
  } 
  \label{fig:examples}
\end{figure}

\paragraph{\JZ{Category retrieval search examples.}} In \figref{fig:examples} we present a search example  when using the $12$-bit and $48$-bit \acro{} from \tblref{tab:cifar10}. Note that increasing the bit rate results in retrieved images that are of the same pose as the query, suggesting that our method has potential for weakly-supervised (automatic) category refinement.

\section{Conclusion}
In this work we introduced \acro{}, a supervised, structured binary code produced by a simple encoding layer compatible with recent deep pipelines. Unlike previous deep binary hash codes, \acro{} features are block-structured, with each block containing a single active bit. We learn our proposed features in a supervised manner by means of a block-wise softmax non-linearity along with two entropy-based penalties. These penalties promote the one-hot quality of the blocks, while encouraging the active bits to employ the available support uniformly. \JZ{While enjoying comparable complexity at fixed bit-rate, \acro{} outperforms the state-of-the art deep hashing methods in the single-domain category retrieval task, as well as state-of-the art structured vector quantizers in the instance retrieval task. \acro{} also outperforms structured vector quantizers in cross-domain category retrieval. Our method further showed promise for weakly-supervised semantic learning, a possible future direction.}


{\small
\bibliographystyle{ieee}
\bibliography{egbib}
}

\end{document}